# BUSIS: A Benchmark for Breast Ultrasound Image Segmentation

Yingtao Zhang[†1], Min Xian[†2], H. D. Cheng*[3], Bryar Shareef[2], Jianrui Ding[4], Fei Xu[3], Kuan Huang[3], Boyu Zhang[5], Chunping Ning[6], Ying Wang[7]

[1]School of Computer Science and Technology, Harbin Institute of Technology, Harbin, China
[2]Department of Computer Science, University of Idaho, Idaho Falls, ID 83401, U.S.A.
[3]Dept. of Computer Science, Utah State University, Logan, UT 84322 U.S.A.
[4]School of Computer Science and Technology, Harbin Institute of Technology, Weihai, China
[5] Institute for Modeling Collaboration and Innovation, University of Idaho, Moscow, ID, U.S.A.
[6] Department of Ultrasound, Affiliated Hospital of Medical College Qingdao University, Qingdao, China
[7] Department of General Surgery, Second Hospital of Hebei Medical University, Shijiazhuang, China
[†] Contributed equally
* Corresponding author

**Abstract**

Breast ultrasound (BUS) image segmentation is challenging and critical for BUS Computer-Aided Diagnosis (CAD) systems. Many BUS segmentation approaches have been studied in the last two decades, but the performances of most approaches have been assessed using relatively small private datasets with different quantitative metrics, which results in a discrepancy in performance comparison. Therefore, there is a pressing need for building a benchmark to compare existing methods using a public dataset objectively, to determine the performance of the best breast tumor segmentation algorithm available today, and to investigate what segmentation strategies are valuable in clinical practice and theoretical study. In this work, a benchmark for B-mode breast ultrasound image segmentation is presented. In the benchmark, 1) we collected **562** breast ultrasound images, prepared a software tool, and involved four radiologists in obtaining accurate annotations through standardized procedures; 2) we extensively compared the performance of **sixteen** state-of-the-art segmentation methods and discussed their advantages and disadvantages; 3) we proposed a set of valuable quantitative metrics to evaluate both semi-automatic and fully automatic segmentation approaches; and 4) the successful segmentation strategies and possible future improvements are discussed in details.

**Keywords:** Breast ultrasound (BUS) images; segmentation; computer-aided diagnosis (CAD); benchmark.

## 1. Introduction

Breast cancer occurs in the highest frequency in women among all cancers and is also one of the leading causes of cancer death worldwide [1]. The key to reducing mortality is to find the signs and symptoms of breast cancer at its early stage. In current clinical practice, breast ultrasound (BUS) imaging with computer-aided diagnosis (CAD) system has become one of the most important and effective approaches for breast cancer detection due to its noninvasive, non-radioactive and cost-effective nature. In addition, it is the most suitable approach for large-scale breast cancer screening and diagnosis in low-resource countries and regions.

CAD systems based on B-mode breast ultrasound (BUS) have been developed to overcome the inter- and intra-variabilities of the radiologists' diagnoses, and have demonstrated the ability to improve the diagnosis performance of breast cancer [2]. Automatic BUS segmentation, extracting tumor region from normal tissue regions of BUS image, is a crucial component in a BUS CAD system. It can change the traditional subjective tumor assessments into operator-independent, reproducible, and accurate tumor region measurements.

Driven by the clinical demand, automatic BUS image segmentation has attracted great attention in the last two decades; and many automatic segmentation algorithms were proposed. The existing approaches can be classified into *semi-automatic* and *fully automatic* according to "with or without" user interactions in the segmentation process. In most semi-automatic methods, the user needs to specify a region of interest (ROI) containing the lesion, a seed in the lesion, or an initial boundary. Fully automatic segmentation is usually considered as a top-down framework that models the knowledge of breast ultrasound and oncology as prior constraints and needs no user intervention at all. However, it is quite challenging to develop automatic tumor segmentation approaches for BUS images, due to the poor image quality caused by speckle noise, low contrast, weak boundary, and artifacts. Furthermore, tumor size, shape, and echo strength vary considerably across patients, which prevents the application of strong priors to object features that are important for conventional segmentation methods.

In previous works, most approaches were evaluated by using private datasets and different quantitative metrics (see Table 1), that make the objective and effective comparisons among the methods quite challenging. As a consequence, it remains difficult to determine the best performance of the algorithms available today, what segmentation strategies are accessible to clinic practice and study, and what image features are helpful and useful in improving segmentation accuracy and robustness.

In this paper, we present a BUS image segmentation benchmark including 562 B-Mode BUS images with ground truths, and compare **sixteen** state-of-the-art BUS segmentation methods by using **seven** popular quantitative metrics. Besides the BUS dataset in this study, three other BUS datasets [66-68] were published recently. [66] and [67] have many challenging images with small tumors and could be valuable to test algorithm performance on segmenting small tumors; but [66] has only 163 images, and [67] does not have ground truths for most images. [68] has 763 images including 133 normal images (without tumors). It is valuable to test algorithms' robustness in dealing with normal images. *However, the three datasets did not use the same standardized process for ground truth generation; therefore, we do not report the performance of the algorithms on them*.

We also make the BUS dataset and the performance of the sixteen approaches available at http://cvprip.cs.usu.edu/busbench. To the authors' best knowledge, this is the first attempt to benchmark the BUS image segmentation methods. With the help of this benchmark, researchers can compare their methods with other algorithms, and find the primary and essential factors for improving the segmentation performance.

The paper is organized as follows: Section 2 gives a brief review of BUS image segmentation; Section 3 illustrates the set-up of the benchmark; in Section 4, the experimental results are presented; and the discussions and conclusion are in Sections 5 and 6, respectively.

Table 1. Recently published approaches.

| Article | Type | Year | Category | Dataset size/Availability | Metrics |
|---|---|---|---|---|---|
| Shareef, et al. [65] | F | 2020 | Deep CNNs | 725/public | TPR, FPR, JI, DSC, AER, AHE, AME |
| Huang, et al. [61] | F | 2018 | Deep CNNs + CRF | 325/private | TPR, FPR, IoU |
| Xian, et al. [3] | F | 2015 | Graph-based | 184/private | TPR, FPR, SI, HD, MD |
| Shao, et al. [18] | F | 2015 | Graph-based | 450/private | TPR, FPR, SI |
| Huang, et al.[4] | S | 2014 | Graph-based | 20/private | ARE,TPVF, FPVF,FNVF |
| Pons, et al [5] | S | 2014 | Deformable models | 163/private | Sensitivity, ROC area |
| Xian, et al. [6] | F | 2014 | Graph-based | 131/private | SI, FPR, AHE |
| Kuo, et al.[7] | S | 2014 | Deformable models | 98/private | DSC |
| Torbati, et al.[8] | S | 2014 | Neural network | 30/private | JI |
| Moon, et al. [9] | S | 2014 | Fuzzy C-means | 148/private | Sensitivity and FP |
| Jiang, et al.[10] | S | 2012 | Adaboost | 112/private | Mean overlap ratio |
| Shan, et al. [11] | F | 2012 | Feedforward neural network | 60/private | TPR, FPR, FNR, HD, MD |
| Yang, et al.[12] | S | 2012 | Naive Bayes classifier | 33/private | FP |
| Shan, et al.[13] | F | 2012 | Neutrosophic L-mean | 122/private | TPR, FPR, FNR, SI, HD, and MD |
| Liu, et al. [14] | S | 2012 | Cellular automata | 205/private | TPR, FPR,FNR, SI |
| Hao, et al. [59] | F | 2012 | DPM + CRF | 480/private | JI |
| Gao, et al. [15] | S | 2012 | Normalized cut | 100/private | TP, FP, SI, HD, MD |
| Hao, et al. [38] | F | 2012 | Hierarchical SVM + CRF | 261/private | JI |
| Liu, et al. [16] | S | 2010 | Level set-based | 79/private | TP, FP, SI |
| Gómez, et al.[17] | S | 2010 | Watershed | 50/private | Overlap ratio, NRV and PD |
| Huang, et al. [51] | F | 2020 | Deep CNNs + CRF | 325/private | TPR, FPR, IoU |
| Huang, et al. [52] | F | 2020 | Deep CNNs | 325/private + 562/public | TPR, FPR, JI, DSC, AER, AHE, AME |

F: fully automatic, S: semi-automatic, SVM: support vector machine, CRF: conditional random field, DPM: deformable part model, CNNs: convolutional neural networks, TP: true positive, FP: false positive, SI: similarity index, HD: Hausdorff distance, MD: mean distance, DSC: Dice similarity, JI: Jaccard index, ROC: Receiver operating characteristic, ARE: average radial error, TPVF : true positive volume fraction , FPVF : false positive volume fraction, FNVF: false negative volume fraction, PR: precision ratio, MR: match rate, NRV: normalized residual value, PD: proportional distance, TPR: true positive ratio, FPR: false positive ratio, FNR: false negative ration, and IoU: Intersection over union.

## 2. Related Works

Many BUS segmentation approaches have been studied in the last two decades, and have proven effective using their datasets. In this section, a brief review of automatic BUS image segmentation approaches is presented. For more details, refer to the survey paper [19]. The BUS image segmentation approaches are classified into five categories: (1) deformable models, (2) graph-based approaches, (3) machine learning-based approaches, (4) classical approaches, and (5) other kinds.

***Deformable models* (*DMs*)**. According to the ways of representing the curves and surfaces, DMs are generally classified into two subcategories: (1) the parametric DMs (PDMs) and (2) the geometric DMs (GDMs). In PDMs-based segmentation approaches, the main work was focused on generating good initial tumor boundaries. [20-24] investigated PDMs by utilizing different preprocessing methods such as the balloon forces, sticks filter, gradient vector flow (GVF) model, watershed approach, etc. In GDMs-based BUS image segmentation approaches, many methods focused on dealing with the weak boundary and inhomogeneity of BUS images. [25-30] utilized the active contour without edges (ACWE) model, Mumford-Shah technique, signal-to-noise ratio and local intensity value, level set approach, phase congruency, etc. [16] proposed a GDMs-based approach which enforced priors of intensity distribution by calculating the probability density difference between the observed intensity distribution and the estimated Rayleigh distribution.

***Graph-based approaches***. Graph-based approaches gain popularity in BUS image segmentation because of the flexibility and efficient energy-optimization. The Markov random field - Maximum a posteriori - Iterated Conditional Mode (**MRF-MAP-ICM**) and the **Graph cuts** are the two major frameworks in graph-based approaches [19]. [62] was a common choice for defining the prior energy [31, 32]. [32-34] utilized Gaussian distribution to model both intensity and texture distributions, and the Gaussian parameters were either from manually selection or from user interaction.

Graph cuts is a special case of the MRF-MAP modeling, but focuses on binary segmentation. [3] proposed a novel fully automatic BUS image segmentation framework in which the graph cuts energy modeled the information from both the frequency and space domains. [35] built the graph using image regions, which was initialized by specifying a group of tumor regions ($F$) and a group of background regions ($B$). [36] were applied to define the weight function of the smoothness term (prior energy). [35] proposed a discriminative graph cut approach in which the data term was determined online by a pre-trained Probabilistic Boosting Tree (PBT) classifier [37]. In [38], a hierarchical multiscale superpixel classification framework was proposed to define the data term.

***Machine learning-based approaches.*** Both supervised and unsupervised learning approaches have been applied to BUS image segmentation. Unsupervised approaches are simple and fast, and commonly utilized as preprocessing to generate candidate image regions. Supervised approaches are good for integrating features at different levels and producing accurate results.

*Clustering*: [39] proposed a BUS image segmentation method by applying the spatial fuzzy c-means (sFCM) [40] to the local texture and intensity features. In [41], FCM was applied to intensities for generating image regions in four clusters. [9] applied FCM to image regions produced by using the mean shift method. [13] extended the FCM and proposed the neutrosophic l-means (NLM) clustering to deal with the weak boundary problem in BUS image segmentation by considering the indeterminacy membership.

*SVM and NN*: [42] trained a support vector machine (SVM) using local image features to categorize small image lattices ($16 \times 16$) into the tumor or non-tumor classes. [10] trained Adaboost classifier using 24 Haar-like features [43] to generate a set of candidate tumor regions. [44] proposed an NN-based method to segment 3D BUS images by processing 2D image slices using local image features. Two artificial neural networks (ANNs) to determine the best-possible threshold were trained [45]. [11] trained an ANN to conduct pixel-level classification by using the joint probability of intensity and texture [20] with two new features: the phase in the max-energy orientation (PMO) and radial distance (RD). The ANN had 6 hidden nodes and 1 output node.

*Deep Learning*: deep learning-based approaches have been reported to achieve state-of-the-art performance for many medical tasks such as prostate segmentation [46], cell tracking [47], muscle perimysium segmentation [48], brain tissue segmentation [49], breast tumor diagnosis [50], etc. [51] combined fuzzy logic with fully convolutional network (FCN), and the 5-layer structure of the breast is utilized to refine the final segmentation results. [52] applied fuzzy logic to five convolutional blocks. It can handle the breast images having no tumors or more than one tumor which could not be processed well before. Deep learning models have great potential to achieve good performance due to the ability to characterize large image variations and to learn compact image representation using a sufficiently huge image dataset *automatically*. Deep learning architectures based on convolutional neural networks (CNNs) were employed in medical image segmentation [46-52]. [60] compared the performance of LeNet [62], UNet [47], and FCN-AlexNet [63] for detecting tumors in BUS images, and the results showed that the patch-based LeNet achieved the best performance for the first dataset (306 images), and the transfer learning-based FCN-AlexNet outperformed other approaches for the second dataset (163 images). [61] utilized fully convolutional CNNs to identify the tissue layers of the breast, and integrated the layer information into a fully connected CRF model to generate the final segmentation results. [65] proposed the STAN architecture to improve the small tumor segmentation. Two encoders were employed in STAN to extract the multi-scale contextual information from different levels of the contracting part.

***Classical approaches:*** Three most popular classical approaches were applied to BUS image segmentation: thresholding, region growing, and watershed. [11] proposed an automatic seed generation approach. [54] defined the cost of growing a region by modeling common contour smoothness and region similarity (mean intensity and size).

Watershed could produce more stable results than thresholding and region growing approaches. [40] selected the markers based on the grey level and connectivity. [56] applied watershed to determine the boundaries of the binary image. The markers were set as the connected dark regions. [58] applied watershed and post-refinement based on the grey level and location to generate candidate tumor regions.

***Other kinds***: Two interesting approaches are in this category: cell computation [35, 36] and cellular automation [14]. *Cell computation:* The cells are the small image regions, and adjacent cells compete with each other to split or merge. [36] defined two types of competitions: Type 1 and Type II. In Type I competition, two adjacent cells from different regions compete to split one cell from a region and merge it into another region. One cell splits from a multi-cell region and generates a single-cell region in Type II competition. This approach is simple and fast, but it needs user interaction to select the tumor region.*Cellular automation (*CA*)*: Each cell in CA has three components: state, neighbors, and a transition function. A cell's state updates by using its transition function and the states of its neighboring cells. [14] constructed the transition function by using local texture correlation. It could generate accurate tumor boundaries and did not have the shrink problem in Graph cuts. The computation cost for CA to reach a stable state set was quite high.

In Table 1, we list 20 BUS image segmentation approaches published recently.

## 3. Benchmark Setup

### 3.1 BUS segmentation approaches and setup

We obtained permissions from the developers of **six** state-of-the-art BUS segmentation methods [3, 11, 14, 16, 18, 65] to use their codes. In addition, we implemented **ten** deep learning-based approaches: Fuzzy FCN [51], Fuzzy FCN Pyramid [52], FCN-AlexNet [63], U-Net [47], SegNet [71], MultiResUNet [72], CE-Net [73], RDAU Net [69], SCAN [70], and DenseU-Net [74]. Approaches in [14] and [16] are interactive, and both need an operator to specify regions of interest (ROIs) manually, and all other approaches are fully automatic.

[16] is a level set-based segmentation approach and sets the initial tumor boundary by user-specified ROI. The maximum number of iterations is set to 450 as the stopping criterion. [14] is based on cellular automation and uses the pixels on the boundary of the ROI specified by the user as the background seeds and pixels on an adaptive cross at the ROI center as the tumor seeds. [11] utilizes a predefined reference point (center of the upper part of the image) for seed generation and pre-trained tumor grey-level distribution for texture feature extraction. We use the same reference point defined in [11] and the predefined grey-level distribution provided by the authors; 10-fold cross-validation is employed to evaluate the overall segmentation performance. [3] and [18] are two graph-based fully automatic approaches. In our experiments, we adopt all the parameters from the original papers correspondingly. [47, 63, 65, 71-74] are deep learning approaches and 5-fold cross-validation was applied to test the performance.

### 3.2 Datasets and Ground Truth Generation

Our BUS image dataset has 562 images among women in ages between 26 to 78 years. The images are collected by the Second Affiliated Hospital of Harbin Medical University, the Affiliated Hospital of Qingdao University, and the Second Hospital of Hebei Medical University using multiple ultrasound devices: GE VIVID 7 and LOGIQ E9, Hitachi EUB-6500, Philips iU22, and Siemens ACUSON S2000. The images from different resources are valuable for testing the robustness of the algorithms. Example images from different devices are shown in Figure 1. Informed consents to the protocol from all patients were acquired. The privacy of the patients is well protected.

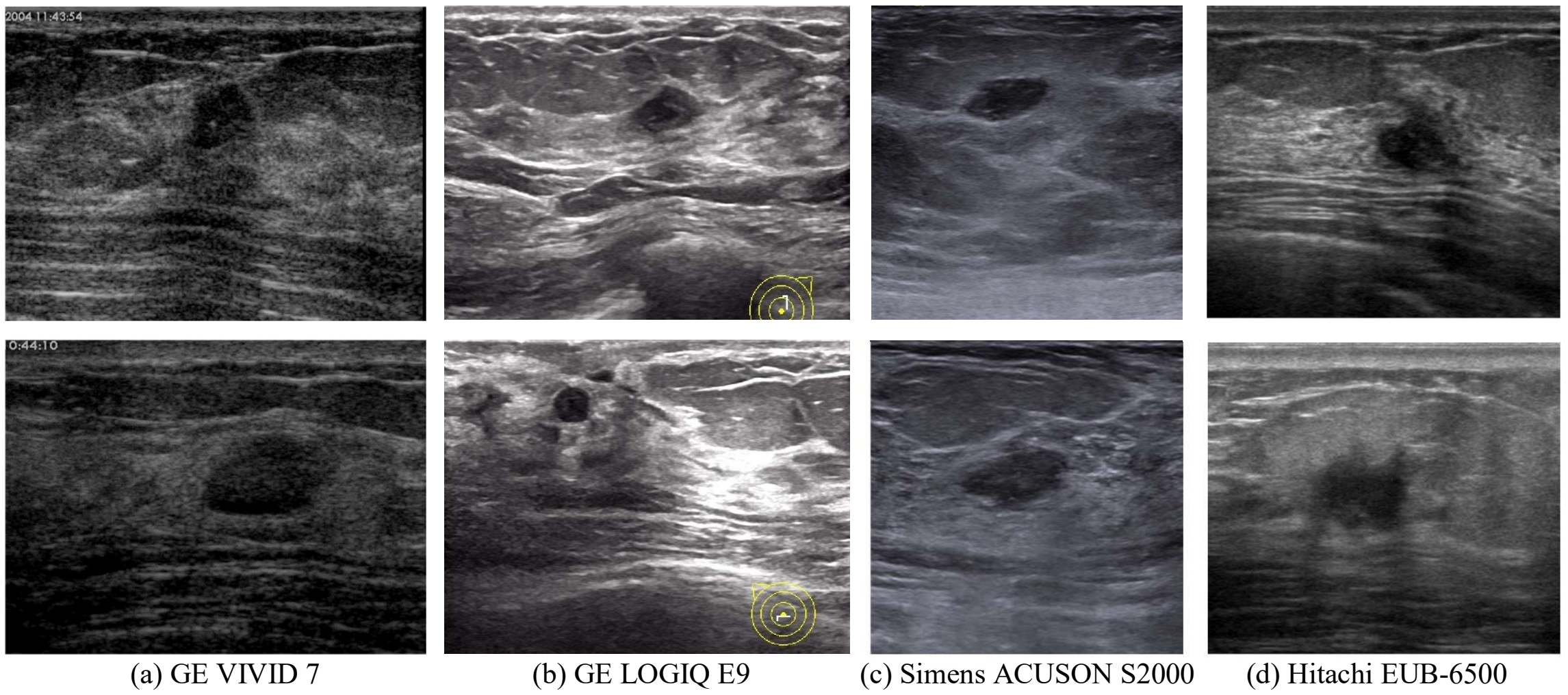

(a) GE VIVID 7 (b) GE LOGIQ E9 (c) Simens ACUSON S2000 (d) Hitachi EUB-6500

Figure 1. Breast ultrasound images collected using different devices.

Four experienced radiologists are involved in the ground truth generation; three radiologists read each image and delineated each tumor boundary individually, and the fourth one (senior expert) will judge if the majority voting results need adjustments. The ground truth generation has four steps: 1) every of the three experienced radiologists delineates each tumor boundary manually, and three delineation results are produced for each BUS image. 2) all pixels inside/on the boundary are viewed as tumor region, outside pixels as background; and majority voting is used to generate the preliminary result for each BUS image. 3) the senior expert read each BUS image and refer to its corresponding preliminary result to decide if it needs any adjustment. 4) We label tumor pixel as **1** and background pixel as **0**; and generate a binary and uncompressed image as the ground truth for each BUS image.

An example of the ground truth generation is in Figure 2.

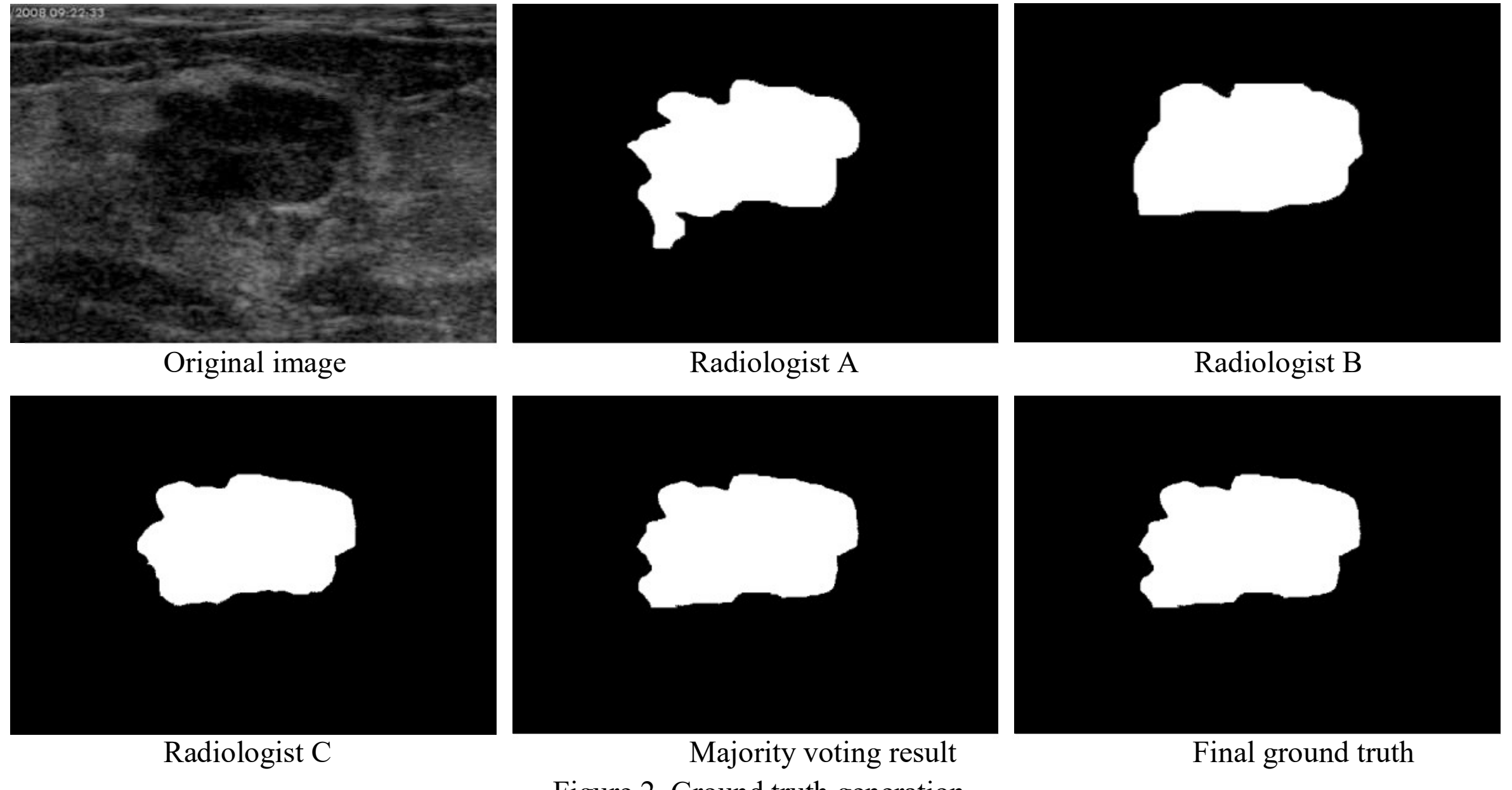

Figure 2. Ground truth generation.

### 3.3 Quantitative metrics

Among the approaches, two of them [14, 16] are semi-automatic, and user predefined region of interest (ROI) needs to be set before the segmentation; while the other fourteen approaches are fully automatic. The performance of semi-automatic approaches may vary with different user interactions. It is meaningless to compare semi-automatic methods with fully automatic methods; therefore, we will compare the methods in two categories separately. In the evaluation of semi-automatic approaches, we compare the segmentation performances of the two methods using the same set of ROIs and evaluate the sensitivity of the methods to ROIs with different looseness ratio ($LR$) defined by

$$LR = \frac{BD}{BD_0}$$

where $BD_0$ is the size of the bounding box of the ground truth and is used as the baseline, and $BD$ is the size of an ROI containing $BD_0$. We produce 10 groups of ROIs with different *LRs* automatically using the approach described in [55]: move the four sides of an ROI toward the image borders to increase the looseness ratio; and the amount of the move is proportional to the margin between the side and the image border. The $LR$ of the first group is 1.1; and the $LR$ of each of the other groups is 0.2 larger than that of its previous group.

The method in [11] is fully automatic, it involves neural network training and testing, and a 10-fold cross-validation strategy is utilized to evaluate its performance. Methods in [3, 18] need no training and operator interaction. All experiments are performed using a windows-based PC equipped with a dual-core (2.6 GHz) processor and 8 GB memory. The performances of these methods are validated by comparing the results with the ground truths. Both area and boundary metrics are employed to assess the performances of the approaches. The area error metrics include the true positive ratio (TPR), false positive ratio (FPR), Jaccard index (JI), Dice's coefficient (DSC), and area error ratio (AER)

$$\mathrm{TPR} = \frac{|A_m \cap A_r|}{|A_m|} \tag{1}$$

$$\mathrm{FPR} = \frac{|A_m \cup A_r - A_m|}{|A_m|} \tag{2}$$

$$\mathrm{JI} = \frac{|A_m \cap A_r|}{|A_m \cup A_r|} \tag{3}$$

$$\mathrm{DSC} = \frac{2|A_m \cap A_r|}{|A_m| + |A_r|} \tag{4}$$

$$\mathrm{AER} = \frac{|A_m \cup A_r| - |A_m \cap A_r|}{|A_m|} \tag{5}$$

where $A_m$ is the pixel set in the tumor region of the ground truth, $A_r$ is the pixel set in the tumor region generated by a segmentation method, and $|\cdot|$ indicates the number of elements of a set. TPR, FPR, and AER take values in [0, 1]; and FPR could be greater than 1 and takes value in [0, $+\infty$). Furthermore, Hausdorf error (HE) and mean absolute error (MAE) are used to measure the worst possible disagreement and the average agreement between two boundaries, respectively. Let $C_m$ and $C_r$ be the boundaries of the tumors in the ground truth and the segmentation result, respectively. The HE is defined by

$$\mathrm{HE}(C_m, C_r) = \max\{\max_{x \in C_m}\{d(x, C_r)\}, \max_{y \in C_r}\{d(y, C_m)\}\} \tag{6}$$

where $x$ and $y$ are the points on the boundaries $C_m$ and $C_r$, respectively; and $d(\cdot, C)$ is the distance between a point and a boundary $C$ as

$$d(z, C) = \min_{k \in C}\{\|z - k\|\}$$

where $\|z - k\|$ is the Euclidean distance between points $z$ and $k$; and $d(z, C)$ is the minimum distance between point $z$ and all points on $C$.

MAE is defined by

$$\mathrm{MAE}(C_m, C_r) = \frac{1}{2\left(\sum_{x \in C_m} \frac{d(x, C_r)}{n_r} + \sum_{y \in C_r} \frac{d(y, C_m)}{n_m}\right)}. \tag{7}$$

In Eq. (7), $n_r$ and $n_m$ are the numbers of points on boundaries $C_r$ and $C_m$, respectively.

The seven metrics above were discussed in [19]. For the first two metrics (TPR and FPR), each of them only measures a certain aspect of the segmentation result, and is not suitable for describing the overall performance; e.g., a high TPR value indicates that most portion of the tumor region is in the segmentation result; however, it cannot claim an accurate segmentation because it does not measure the ratio of correctly segmented non-tumor regions. The other five metrics (JI, DSC, AER, HE and MAE) are more comprehensive and effective to measure the overall performance of the segmentation approaches and are commonly applied to tune the parameters of the segmentation models [3], e.g., large JI and DSC and small AER, HE and MAE values indicate the high overall segmentation performance.

Although JI, DSC, AER, HE, and MAE are comprehensive metrics, we still recommend using both TPR and FPR for evaluating BUS image segmentation; since with these two metrics, we can discover some hidden characteristics that cannot be found through the comprehensive metrics. Suppose that the algorithm has low overall performance (small JI and DSC, and large AER, HE and MAE); if FPR and TPR are large, we can conclude that the algorithm has overestimated the tumor region; if both FPR and TPR are small, the algorithm has underestimated the tumor regions. The findings from TPR and FPR can guide the improvement of the algorithms.

## 4. Approaches Comparison

In this section, we evaluate 16 state-of-the-art approaches [3, 11, 14, 16, 18, 47, 51, 52, 63, 65, 69-74]. The 16 approaches are selected based on three criteria: (1) select at least one representative approach for each category except the classic approaches; (2) each approach should achieve good performance in their original dataset; and (3) the source codes of the approaches are available. [3, 11, 18, 47, 51, 52. 63, 65, 69-74] are fully automatic, and we compare their average performances by using the seven metrics discussed in section III (C); while for semi-automatic approaches [14, 16], we also evaluate their sensitivities of the seven metrics with different *LRs*.

### 4.1 Semi-automatic segmentation approaches

Ten ROIs have been generated automatically for each BUS image, and *LRs* range from 1.1 to 2.9 (step size is 0.2). Totally, 5620 ROIs are generated for the entire BUS dataset, and we run each of the semi-automatic segmentation approach 5620 times to produce the results. All the segmentation results on the ROIs with the same *LR* are utilized to calculate the average TPR, FPR, DSC, AER, HE, and MAE, respectively; and the results of [14] and [16] are shown in Figures 3 and 4, respectively.

The segmentation results of [14] are demonstrated in Figure 3. All average JI values are between 0.7 and 0.8; and all average DSC values are between 0.8 and 0.9. The average TPR values are above 0.7, and increase with *LR*s of ROIs; the average JI and DSC values increase firstly, and then decrease; the average FPR values increase with the increasing looseness of ROIs; and the average DSC, HE and MAE decrease firstly, and then increase. Five metrics (average JI, DSC, AER, HE and MAE) reach their optimal values at the *LR* of 1.9 (Table 2).

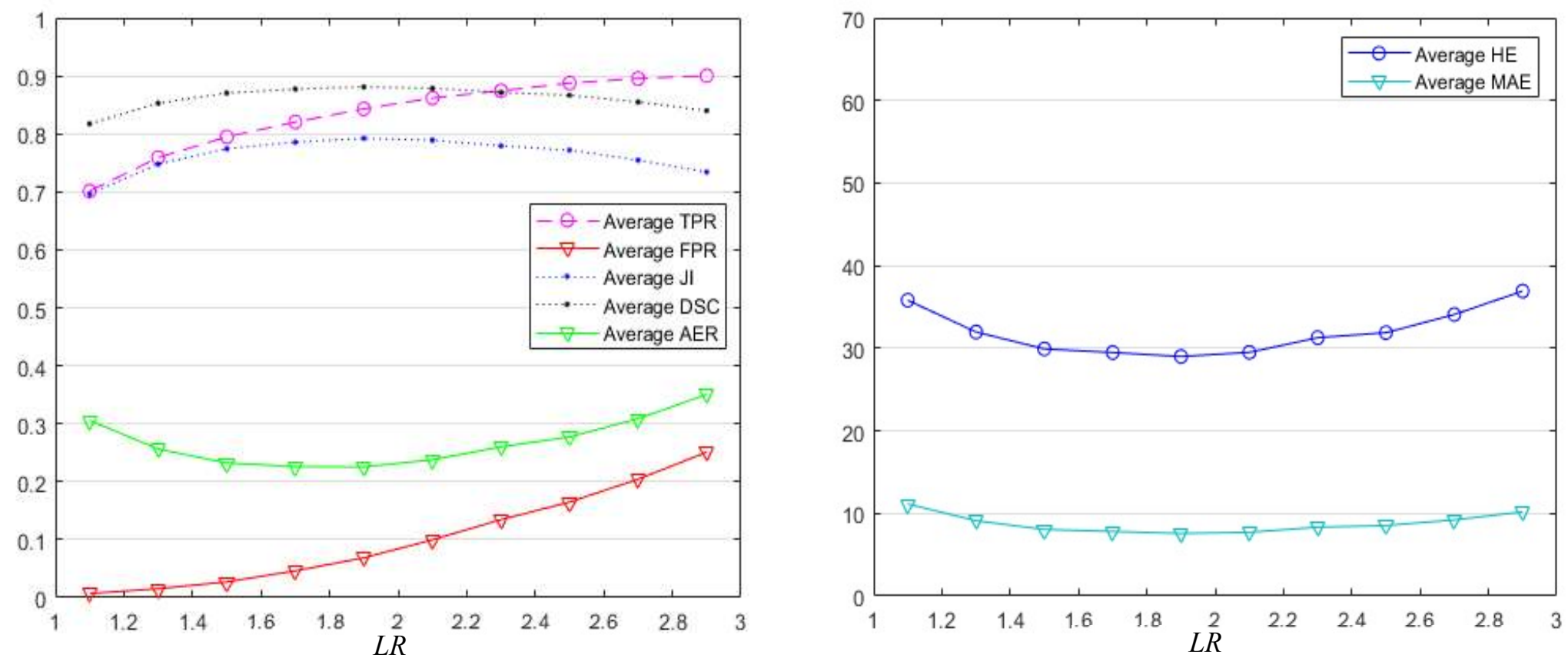

Figure 3. Average segmentation results of [14] using ROIs with diferent looseness ratio (LR).

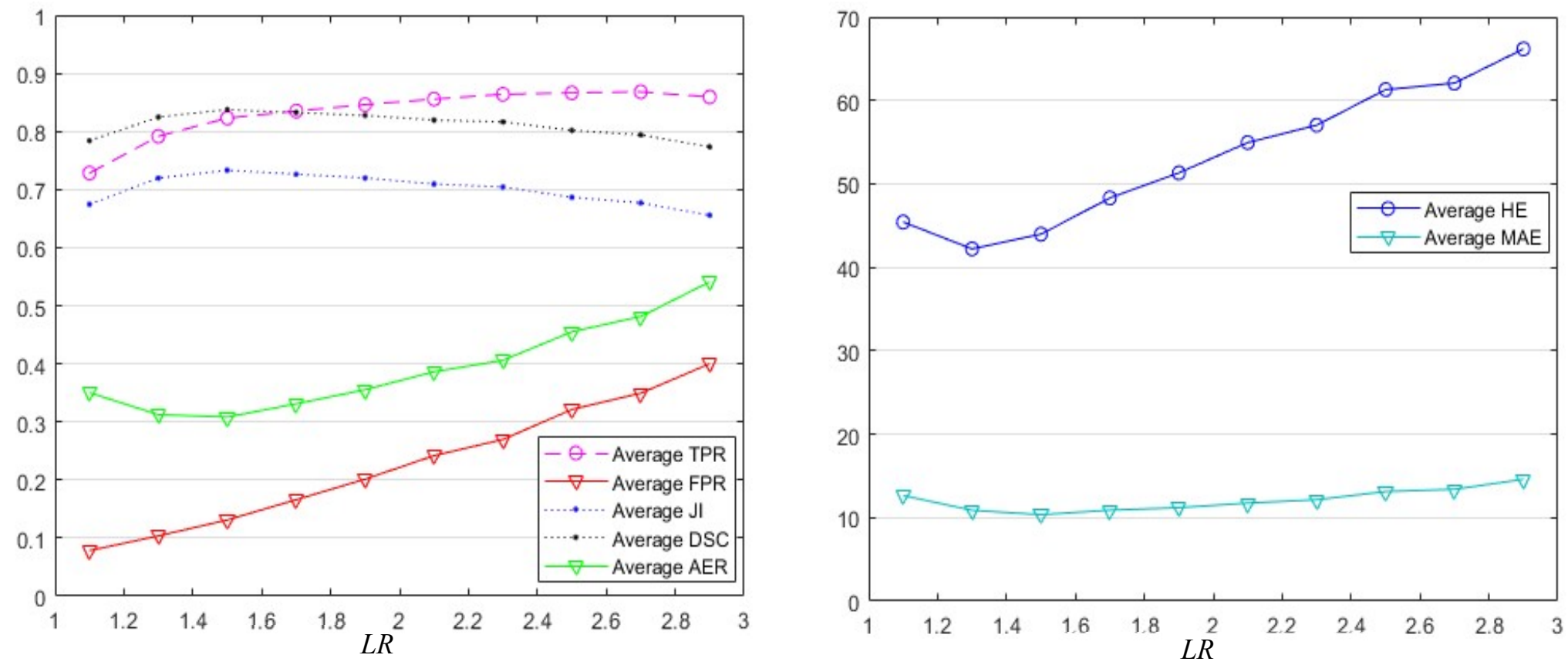

Figure 4. Average segmetation results of [16] using ROIs with different looseness ratio (LR).

As shown in Figure 4, all the average TPR and DSC values of the method in [16] are above 0.7, and the average JI values vary in the range [0.65, 0.75]. The average TPR values increase with the increasing *LR* values of ROIs. Both the average JI and DSC values tend to increase firstly, and then decrease with the increasing *LR*s of ROIs. FPR, AER and HE have low average values when the *LR*s are small, which indicate that the high performance of the method in [16] can be achieved by using tight ROIs; however, the values of the three metrics increase almost linearly with the *LR*s of ROIs when the looseness is greater than 1.3; this observation shows that the overall performance of [16]

Table 2. Quantitative results of [14] and [16] using 10 *LR*s of ROI.

| Metrics / Methods | *Looseness Ratio* | **Area error metrics** | | | | | **Boundary error metrics** | | **Time** |
|---|---|---|---|---|---|---|---|---|---|
| | | Ave. TPR | Ave. FPR | Ave. JI | Ave. DSC | Ave. AER | Ave. HE | Ave. MAE | Ave. Time (s) |
| [16] | 1.1 | 0.73(0.23) | **0.08**(0.09) | 0.67(0.20) | 0.78(0.18) | 0.35(0.22) | 45.4(31.6) | 12.6(10.9) | 18 |
| | 1.3 | 0.79(0.18) | 0.10(0.12) | 0.72(0.16) | 0.82(0.14) | 0.31(0.19) | **42.2(28.0)** | 10.9(8.9) | 22 |
| | **1.5** | 0.82(0.15) | 0.13(0.14) | **0.73**(0.14) | **0.84**(0.11) | **0.31(0.18)** | 44.0(28.3) | **10.4(7.5)** | 27 |
| | 1.7 | 0.83(0.15) | 0.17(0.18) | **0.73**(0.14) | 0.83(0.12) | 0.33(0.20) | 48.3(32.2) | 10.9(8.0) | 27 |
| | 1.9 | 0.85(0.14) | 0.20(0.21) | 0.72(0.14) | 0.83(0.12) | 0.36(0.23) | 51.3(35.3) | 11.2(7.9) | 30 |
| | 2.1 | 0.86(0.14) | 0.24(0.25) | 0.71(0.15) | 0.82(0.13) | 0.39(0.27) | 54.9(38.8) | 11.7(8.4) | 30 |
| | 2.3 | 0.86(0.13) | 0.27(0.28) | 0.70(0.15) | 0.82(0.12) | 0.41(0.29) | 57.0(41.7) | 12.1(8.8) | 36 |
| | 2.5 | **0.87**(0.14) | 0.32(0.33) | 0.69(0.16) | 0.80(0.13) | 0.46(0.34) | 61.3(44.2) | 13.1(10.5) | 39 |
| | 2.7 | **0.87**(0.14) | 0.35(0.36) | 0.68(0.17) | 0.79(0.14) | 0.48(0.36) | 62.1(43.3) | 13.4(9.5) | 40 |
| | 2.9 | 0.86(0.17) | 0.40(0.41) | 0.66(0.19) | 0.77(0.17) | 0.54(0.44) | 66.2(46.1) | 14.6(10.7) | 44 |
| [14] | 1.1 | 0.70(0.10) | **0.01**(0.02) | 0.70(0.09) | 0.82(0.07) | 0.31(0.09) | 35.8(17.0) | 11.1(5.3) | 487 |
| | 1.3 | 0.76(0.09) | 0.02(0.03) | 0.75(0.08) | 0.85(0.06) | 0.26(0.09) | 32.0(15.6) | 9.1(4.6) | 467 |
| | 1.5 | 0.79(0.08) | 0.03(0.04) | 0.77(0.08) | 0.87(0.05) | **0.23**(0.09) | 29.9(15.0) | 8.1(4.2) | 351 |
| | 1.7 | 0.82(0.09) | 0.05(0.06) | **0.79**(0.09) | **0.88**(0.06) | **0.23**(0.10) | 29.5(16.5) | 7.8(4.8) | 341 |
| | **1.9** | 0.84(0.09) | 0.07(0.07) | **0.79**(0.09) | **0.88**(0.06) | **0.23**(0.11) | **29.0**(17.0) | **7.6**(5.3) | 336 |
| | 2.1 | 0.86(0.08) | 0.10(0.09) | **0.79**(0.10) | **0.88**(0.07) | 0.24(0.13) | 29.5(18.4) | 7.7(5.2) | 371 |
| | 2.3 | 0.87(0.09) | 0.13(0.12) | 0.78(0.11) | 0.87(0.08) | 0.26(0.16) | 31.3(21.9) | 8.3(6.4) | 343 |
| | 2.5 | 0.89(0.09) | 0.16(0.14) | 0.77(0.11) | 0.87(0.08) | 0.28(0.17) | 31.9(20.1) | 8.5(6.1) | 365 |
| | 2.7 | **0.90**(0.09) | 0.20(0.15) | 0.75(0.11) | 0.85(0.08) | 0.31(0.18) | 34.1(20.2) | 9.2(5.9) | 343 |
| | 2.9 | **0.90**(0.10) | 0.25(0.18) | 0.73(0.12) | 0.84(0.10) | 0.35(0.22) | 36.9(21.8) | 10.2(6.7) | 388 |

*The values in '( )'are the standard deviations.

drops rapidly by using large ROIs above a certain level of *LR*. The average MAE values decrease firstly, and then increase and vary with the *LRs* in a small range. Four metrics (average JI, DSC, AER and MAE) reach their optimal values at the *LR* of 1.5 (Table II). After 1.5, the increasing ROIs make [16] segment more non-tumor regions into the tumor region (refer to the average FPR curve in Figure 4). The increasing false positive results in the decreasing of the average, JI and DSC values, and increasing of all other metrics.

As shown in Figures 3 and 4, and in Table 2, the two approaches achieve their best performances with different *LRs* (1.5 and 1.9 respectively). We can observe the following facts:

- [14] and [16] are quite sensitive to the sizes of ROIs.
- [14] and [16] achieve the best performance when setting them with their optimal *LR*s (1.9 for [14] and 1.5 for [16]).
- The performances of the two approaches drop if the looseness level is greater than a certain value; and the performance of the method [14] drops much slower than that of the method in [16].

Table 3. Overal Performance of All approaches.

| Metrics / Methods | **Area error metrics** | | | | | **Boundary error metrics** | |
|---|---|---|---|---|---|---|---|
| | Ave. TPR | Ave. FPR | Ave. JI | Ave. DSC | Ave. AER | Ave. HE | Ave. MAE |
| FCN-AlexNet [61] | **0.95**/-- | 0.34/-- | 0.74/-- | 0.84/-- | 0.39/-- | 25.1/-- | 7.1/-- |
| SegNet [69] | 0.94/-- | 0.16/-- | 0.82/-- | 0.89/-- | 0.22/-- | 21.7/-- | 4.5/-- |
| U-Net [47] | 0.92/-- | 0.14/-- | 0.83/-- | 0.90/-- | 0.22/-- | 26.8/-- | 4.9/-- |
| CE-Net [71] | 0.91/-- | 0.13/-- | 0.83/-- | 0.90/-- | 0.22/-- | 21.6/-- | 4.5/-- |
| MultiResUNet [70] | 0.93/-- | 0.11/-- | 0.84/-- | 0.91/-- | 0.19/-- | **18.8**/-- | 4.1/-- |
| RDAU NET [67] | 0.91/-- | 0.11/-- | 0.84/-- | 0.91/-- | 0.19/-- | 19.3/-- | 4.1/-- |
| SCAN [68] | 0.91/-- | 0.11/-- | 0.83/-- | 0.90/-- | 0.20/-- | 26.9/-- | 4.9/-- |
| DenseU-Net [72] | 0.91/-- | 0.16/-- | 0.81/-- | 0.88/-- | 0.25/-- | 25.3/-- | 5.5/-- |
| STAN [63] | 0.92/-- | 0.09/-- | 0.85/-- | 0.91/-- | 0.18/-- | 18.9/-- | **3.9**/-- |
| Xian, et al.[3] | 0.81/0.91 | 0.16/0.10 | 0.72/0.84 | 0.83/-- | 0.36/-- | 49.2/24.4 | 12.7/5.8 |
| Shan, et al. [11] | 0.81/0.93 | 1.06/0.13 | 0.60/-- | 0.70/-- | 1.25/-- | 107.6/18.9 | 26.6/5.0 |
| Shao, et al. [18] | 0.67/0.81 | 0.18/0.12 | 0.61/0.74 | 0.71/-- | 0.51/-- | 69.2/50.2 | 21.3/13.4 |
| Huang, et al. [51] | 0.94/-- | 0.08/-- | **0.88**/-- | 0.92/-- | **0.14**/-- | 19.8/-- | 4.2/-- |
| Huang, et al. [52] | 0.93/0.93 | **0.07**/0.07 | 0.87/0.87 | **0.93**/0.93 | 0.15/0.15 | 26.0/26.0 | 4.9/4.9 |
| Liu, et al. [16] *LR* = 1.5 | 0.82/0.94 | 0.13/0.08 | 0.73/0.87 | 0.84/-- | 0.31/-- | 44.0/26.3 | 10.4/-- |
| Liu, et al. [14] *LR* = 1.9 | 0.84/0.94 | **0.07**/0.07 | 0.79/0.88 | 0.88/-- | 0.23/-- | 29.0/25.1 | 7.6/-- |

*The values before the slashes are approaches' performances on the proposed dataset, and after the slashes are their performances reported in the original publications. Notation '--' indicates that the corresponding metric was not reported in the original paper.

- Set 1.9 as the optimal *LR* for [14] and 1.5 for [16]; and [14] achieves better average performance than that of [16].
- The running time of the approach in [16] is proportional to the size of specified ROI, while there is no such relationship of the running time for the approach in [14].
- The running time of the approach in [14] is slower than that of the approach in [16] by one order of the magnitude.

4.2 Fully automatic segmentation approaches

The performance of **14** fully automatic approaches is reported in Table 3. Except for methods [3], [11], and [18], the other approaches are deep convolutional neural networks. In general, all deep learning approaches outperform [3], [11], and [18] using the benchmark dataset. [3] achieves better performance than that of the methods in [11] and [18] on all five comprehensive metrics. [14] and [52] achieve the lowest average FPR. The method in [11] has the same average TPR value as the method in [3]; however, its average FPR value is much high (1.06) which is almost six times larger than that of the method in [3]; the high average FPR and AER values of the method in [11] indicate that large portions of non-tumor regions are misclassified as tumor regions. The average JIs of all deep learning approaches are above 0.8 except FCN-AlexNet; and [51] achieved the best average JI performance. Table 3 also shows the average optimal performances of [16] and [14] at the *LR*s of 1.5 and 1.9, respectively.

**5. Discussions**

Many semi-automatic segmentation approaches are utilized for BUS image segmentation [19]. User interactions (setting seeds and/or ROIs) are required by these approaches and could be useful for segmenting BUS images with extremely low quality. As shown in Table III, the two interactive approaches could achieve very good performance if the ROI is set properly.

Figures 3 and 4 also demonstrate that the two semi-automatic approaches achieve varying performances using different sizes of ROIs. Therefore*, the major issue in semi-automatic approaches is to determine the best ROIs/seeds. But such issue has been neglected before completely*. Most semi-automatic approaches focused only on improving segmentation performance by designing complex features and segmentation models, but failed to consider user interaction as an important factor that could affect the segmentation performance. Hence, we recommend researchers that they should consider such issues when they develop semi-automatic approaches. *Two possible solutions could be employed to solve this issue. First, for a given approach, we could choose the best LR by running experiments on a given BUS image training set (like Section 4.1) and apply the LR to the test set. Second, like the interactive segmentation approach in [57], we could bypass this issue by designing segmentation models less sensitive to user interactions*.

Fully automatic segmentation approaches have many good properties such as operator-independence and reproducibility. *The key strategy that shared by many successful fully automatic approaches is to localize the tumor ROI accurately by modeling domain knowledge*. [11] localizes tumor ROI by formalizing the empirical tumor location, appearance, and size; [3] generates tumor ROI by finding adaptive reference position; and in [18], the ROI is generated to detect the mammary layer of BUS image, and the segmentation algorithm only detects the tumor in this layer. However, in many fully automatic approaches, the performance heavily depends on hand-crafted features and some inflexible constraints, e.g., [11] utilizes a fixed reference position to rank the candidate regions in the ROI localization process. Table 3 demonstrates that deep Learning approaches outperform all traditional approaches. It is worth noting that deep learning approaches have limitations in segmenting small breast tumors [65].

As shown in Table 3, using the benchmark dataset, approaches [3, 11, 14, 16, 18] cannot achieve the performances reported in the original papers. The average JI of [3] is 14% less than the original average JI; the average FPR of [11] is 87% higher than the original value; the average TPR of [18] is 17% less than its reported value; and the average JI values of [14] and [16] are 17% and 10% lower than the reported values, respectively. The major reasons are: (1) the large image varieties of the benchmark dataset; and (2) the lack of robustness of the approaches when dealing with images from different resources.

As shown in Table 1, many quantitative metrics exist for evaluating the performances of BUS image segmentation approaches. In this paper, we have applied seven metrics [19] to evaluate BUS image segmentation approaches. As shown in Figures 3 and 4, average JI, DSC, and AER have the same trend, and each of them is sufficient to evaluate the area error comprehensively.

## 6. Conclusion

In this paper, we establish a BUS image segmentation benchmark and present the comparing results of **16** state-of-the-art segmentation approaches; two of them are semi-automatic and others are fully automatic. The BUS dataset contains **562** BUS images collected using three different ultrasound machines; therefore, the images have a large variance in terms of image contrast, brightness and level of noise, and can be valuable for testing the robustness of the algorithms as well. In the approaches, two of them [3, 18] are graph-based approaches, [16] is a level set-based segmentation approach, [11] is ANN-based approach, [14] is based on cell competition, and [47, 51, 52, 65, 67, 71-76] are deep convolutional neural networks.

The quantitative analysis of the considered approaches highlights the following findings.

- As shown in Table 3, by using the benchmark, no approaches in this study can achieve the same performances reported in their original papers.
- The two semi-automatic approaches are quite sensitive to user interaction (*LR*). See Figures 3 and 4.
- Deep learning approaches outperform all conventional approaches using our benchmark dataset.
- The quantitative metrics such as JI, DSC, AER, HE, and MAE are more comprehensive and effective to measure the overall segmentation performance than TPR and FPR; however, TPR and FPR are also useful for developing and improving algorithms.

In addition, the benchmark should be and will be expanded continuously.

## 7. Acknowledgement

This work was supported, in part, by the Institute for Modeling Collaboration (IMCI) at the University of Idaho through NIH Award #P20GM104420. This work was also supported, in part, by the Chinese NSF (81501477) and by the Livelihood Technology Project of Qingdao city (15-9-2-89-NSH). We would like to acknowledge Dr. Shuangquan Jiang and Dr. Ping Xing for labeling the images, and acknowledge Dr. Juan Shan, Dr. Bo Liu, Dr. Yan Liu and Mr. Haoyang Shao for providing the source codes of their published BUS image segmentation approaches. We also acknowledge Dr. Jiawei Tian for providing 220 images for the benchmark dataset.